\documentclass{article}

\usepackage{arxiv}

\usepackage[utf8]{inputenc} 
\usepackage[T1]{fontenc}    
\usepackage{hyperref}       
\usepackage{url}            
\usepackage{booktabs}       
\usepackage{amsfonts}       
\usepackage{nicefrac}       
\usepackage{microtype}      

\usepackage{times}
\usepackage{epsfig}
\usepackage{graphicx}
\usepackage{amsmath}
\usepackage{amssymb}
\usepackage{balance}

\newcommand{\ma}[1] {\ensuremath{\mathbf{#1}}}

\newcommand{\vectorNorm}[1]{\left|\left|#1\right|\right|}

\newcommand{\Tr}[1]{\ensuremath{\mathrm{tr}({#1}) } }

\newcommand{\Prs}{\ensuremath{\mathbb{P}} }
\renewcommand{\Pr}[1]{\Prs(#1)}

\newcommand{\Ep}[1]{\ensuremath{\mathbb{E} \left[ #1 \right]}}
\newcommand{\Epp}[2]{\ensuremath{\mathbb{E}_{#1}\left[ #2 \right]}}

\newcommand{\KLD}[2]{\mathrm{D_{KL}}({#1}||{#2})}

\usepackage{float}
\usepackage{caption}
\usepackage{subcaption}
\usepackage{booktabs}
\usepackage{multirow}
\usepackage{bm}
\usepackage{enumitem}
\newlist{step}{enumerate}{1}
\setlist[step]{label=Step~\arabic*:}
\newcommand{\IS}{\textrm{IS}}
\usepackage{multicol}

\DeclareUnicodeCharacter{20AC}{ }

\newcommand{\name}{AGAN}

\title{\name: Towards Automated Design of Generative Adversarial Networks}

\author{
  Hanchao Wang, Jun Huan \\
  Big Data Lab, Baidu Research \\
  \texttt{\{wanghanchao01, huanjun\}@baidu.com} \\
}

\begin{document}
\maketitle
\begin{abstract}
   Recent progress in Generative Adversarial Networks (GANs) has shown promising signs of improving GAN training via architectural change. Despite some early success, at present the design of GAN architectures requires human expertise, laborious trial-and-error testings, and often draws inspiration from its image classification counterpart. In the current paper, we present the first neural architecture search algorithm, automated neural architecture search for deep generative models, or \name~for abbreviation, that is specifically suited for GAN training. For unsupervised image generation tasks on CIFAR-10, our algorithm finds architecture that outperforms state-of-the-art models under same regularization techniques. For supervised tasks, the automatically searched architectures also achieve highly competitive performance, outperforming best human-invented architectures at resolution $32\times32$. Moreover, we empirically demonstrate that the modules learned by \name~are transferable to other image generation tasks such as STL-10.
\end{abstract}


\section{Introduction}

Generative Adversarial Networks (GANs) have attracted much research interest since its introduction \cite{GAN} with a wide range of applications, such as image generation, text-to-image synthesis, style transfer \cite{super-resolution,text-to-image,CycleGAN} among many others. GAN learns a target distribution by involving two deep neural networks, namely the generator G and the discriminator D, in a minimax game. The generator G aims to generate samples that resemble real samples from the target distribution, while the discriminator D aims to distinguish the generated samples from the real samples. The model is then trained with simultaneous SGD until a Nash equilibrium is achieved.

Due to the process of minimax optimization and simultaneous SGD, GAN is known to suffer from training instabilities. To mitigate the issue, a string of works focus on the choice of GAN objective function. Notably, in Wasserstein GAN \cite{WGAN}, the authors propose to minimize the Wasserstein distance between the model and target distributions, instead of the original Jensen-Shannon divergence. In LS-GAN \cite{LS-GAN}, the authors consider a least square loss which corresponds to minimizing the Pearson $\chi^2$ divergence between the distributions. In $f$-GAN \cite{f-GAN}, the authors show that any $f$-divergence can be used for GAN objective. Another line of works focus on regularization and normalization techniques, especially the Lipschitz continuity of the discriminator and the conditioning of the generator \cite{generator-conditioning}. Prominent examples include gradient penalty \cite{WGAN-GP} which penalizes the model when the gradient norm moves away from $1$, and spectral normalization \cite{SNGAN} which normalizes the largest singular value by layer using power iteration method. 

Different from existing approaches, we investigate the direction of automating the design of neural architectures to stabilize GAN training and improve performance. There are empirical evidences \cite{WGAN-GP,SNGAN} suggesting that generator and discriminator architectures may have impacts on the stability of GAN training, and hence quality and diversity of images generated by GAN. Despite those early evidences, we observe that DCGAN-style \cite{DCGAN} and ResNet \cite{WGAN-GP} architectures are by far the most prevailing architectures in the GAN literature. Such architectures are built upon highly successful modules used primarily in discriminative tasks, and their optimality in generative model construction is questionable. 



 

Neural architecture search (NAS) has emerged as a promising research direction in recent years. On benchmark data sets including Penn Treebank, CIFAR-10 and ImageNet, NAS algorithms are proven to be capable of designing architectures that rival or even outperform the best human-invented architectures \cite{NAS1,NAS2,regularized-evolution}. The direct application of NAS to GAN architecture design is, however, non-trivial, due to at least two factors. First, the generator of GAN consists of up-sampling modules which are almost never used in any image classifications. Typical image classifications only use down-sampling modules and hence we could not borrow experience from well-studied NAS directly.  Second, architectures of GAN have been much less explored. Comparing to traditional NAS application of image classification we hereby aim at searching through a large variety of topological structures with less human prior knowledge imposed. To the best of our knowledge, we are the first group that aims to perform automated architecture design of deep generative models. 

In order to design an automated neural architecture search, we used reinforcement learning. In our algorithm we used an RNN module to encode the architectures for the up-sampling, down-sampling, and normal modules in GAN. We carefully crafted the search space and proposed a new form of reward shaping functions so that the algorithm is guided faster towards promising architectures. We have performed comprehensive experimental study to evaluate architecture novelty, their performance,  and the transferability of the identified GAN architectures. 


In sum, our main contributions are described below. 
\begin{itemize}
    \item We presented the first automated neural architecture search algorithm, \name, that is specifically designed for the optimization of neural network architectures in deep generative models.
    \item We have identified $3$ novel, modularized architectures, \name-A, \name-B, and \name-C with distinct architectures. 
    \item In our comprehensive experimental study we found that \name-A, \name-B, and \name-C have comparable performance to the best GAN models designed by human-experts. In addition \name-C outperforms the state-of-the-art models under same regularization techniques for unsupervised image generation tasks on CIFAR-10. 
    \item We empirically evaluated and confirmed that the modules learned by \name~are transferable to other data sets such as STL-10. 
\end{itemize}

The rest of the paper is organized with the following sections. In Section \ref{sec:relatedWork}, we present an overview of GAN and NAS. We discuss our methodology in Section \ref{sec:method} and present our experimental evaluation of \name~in Section \ref{sec:experimentalStudy}. In Section \ref{sec:conclusion} we provide a brief discussion of the differences that we observed between our search and traditional NAS search and conclude there. 

\section{Related Work}{\label{sec:relatedWork}}
Equipped with multilayer perceptrons as generator and discriminator, the original GAN \cite{GAN} can successfully learn the data distribution of MNIST, but fails at more complicated image generation tasks.
In DCGAN \cite{DCGAN}, the authors propose a novel class of CNNs as generator and discriminator, together with a set of architecture guidelines for stable convolutional GAN training. Most notably, in generator, the spatial activation size is doubled every layer while the number of output channels is halved; the discriminator much resembles the reverse of  generator. Gulrajani et al. \cite{WGAN-GP} propose a ResNet architecture for GAN on CIFAR-10. In particular, the residual blocks in the generator perform nearest-neighbor up-sampling before the second convolution while some blocks in the discriminator perform average pooling after the second convolution. Many later GAN models are built upon DCGAN-style or ResNet architecture, such as SNGAN \cite{SNGAN}, SAGAN \cite{SAGAN} and BigGAN \cite{BigGAN}. In SAGAN, the authors propose a self-attention layer that models the non-local dependency between high-resolution and low-resolution feature maps. They also make a minor modification of the discriminator by altering the number of hidden layer output channels in residual blocks. In BigGAN, the authors introduce further architectural change including shared class embedding and skip connections in latent variable $z$.

Concerning the architecture design of GAN, a particular line of works focus on how to make use of label information to improve the performance of GAN. In classic conditional GAN \cite{cGAN} framework, label information is concatenated to the input or hidden representations to model a conditional distribution. Miyato et al. \cite{proj-dis} propose to use projection based way to incorporate label information into the discriminator. On the other hand, De Vries et al. \cite{cond-batch-norm} introduce Conditional Batch Normalization to visual question answering tasks, which learns a scale and a shift for each class label. Conditional Batch Norm is widely used in image generation models \cite{SNGAN,SAGAN,BigGAN} to provide label information for the generator of GAN.

Neural architecture search algorithm requires an objective, quantitative metric to measure the performance of the underlying models. In case of GAN, well-studied methods such as kernel density estimation (KDE, or Parzen window estimation) have been questioned as a suitable indicator of visual fidelity of generated images \cite{Theis}. Inception Score (IS) \cite{Inception-score} and Frechet Inception Distance (FID) are arguably the most popular evaluation metrics in the literature. IS uses a pre-trained Google Inception model  \cite{Google-Inception} to classify generated samples. It is defined as
\[
\textrm{IS}(\Prs_g) = \exp{\left( \Epp{x\sim \Prs_g}{\KLD{p(y|x)}{p(y}} \right)},
\]
where $\Prs_g$ is the generated distribution, $p(y|x)$ is the conditional label distribution through the Inception model, and $p(y)$ is marginal of $p(y|x)$ over $\Prs_g$. Similarly FID uses Google Inception model as a feature extractor and computes the distance between the real distribution $\Prs_r$ and $\Prs_g$ as
\[
\textrm{FID}(\Prs_r,\Prs_g) = \vectorNorm{\mu_r-\mu_g}
                            + \Tr{\ma{C}_r + \ma{C}_g - 2(\ma{C}_r\ma{C}_g)^{1/2}},
\]
where $\mu_r$, $\mu_g$, $\ma{C}_r$, $\ma{C}_g$ are the mean and covariance of the real and generated distributions of the extracted features.

NAS became a mainstream research topic since Zoph and Le \cite{NAS1} found state-of-the-art recurrent cell on Penn Treebank and highly competitive architecture on CIFAR-10 using Reinforcement Learning (RL). Various RL methods have been successfully applied to NAS including vanilla policy gradient \cite{EAS,path-level-EAS}, Proximal Policy Optimization (PPO) \cite{NAS2,ENAS} and Q-learning \cite{MetaQNN,BlockQNN}. An alternative approach is to use evolution algorithm \cite{evolution,regularized-evolution,hierarchical-CNN}, maintaining and evolving a large population of neural architectures. In contrast to aforementioned gradient-free optimization methods, Liu et al. \cite{DARTS} propose a gradient-bases search strategy based on continuous relaxation of architecture representation. Other gradient-based approaches include Neural Architecture Optimization (NAO) \cite{NAO} and ProxylessNAS \cite{proxyless-NAS}.

Inspired by Google Inception model, Zoph et al. \cite{NAS2} and Zhong \cite{BlockQNN} propose a search space based on two types of convolutional cells, named normal and reduction cell. This design leads to a simplified yet quality search space and enables the transferability of resulting architecture found by NAS. It is widely adopted by many later works \cite{PNAS,ENAS,regularized-evolution,DARTS,NAO}. Our work also falls in the category of searching cell topology but differs in the following ways:
\begin{itemize}
    \item In all previous RL-based NAS algorithms, the convolutional cell solely consists of unitary and binary operations except for the final concatenation. In another word, candidate cell topology can only be DAG with indegree no greater than $2$. Our architecture representation allows searching through cells with arbitrary topology.
    \item Previous works search for discriminative models with normal or down-sampling modules, we search for generative models where up-sampling modules play a significant role.
\end{itemize}

\section{Method} {\label{sec:method}}

Our work makes use of the Neural Architecture Search with Reinforcement Learning framework proposed by \cite{NAS1}. A controller recurrent neural network (RNN) samples architectures of the generator and discriminator of GAN simultaneously. The sampled architectures are then sent to computation nodes for training and evaluation using Inception Score. The resulting performance is used as feedback to update controller RNN parameters using REINFORCE rule \cite{reinforce}. Below we provide detailed description of the three critical components in our design: (i) controller architectures, (ii) the set of operations that we use to construct a GAN (a.k.a. the search space), and (iii) how to train a reinforcement learning. 

\subsection{Controller architecture}

\begin{figure*}[h!]
    \centering
    \includegraphics[width=.8\textwidth]{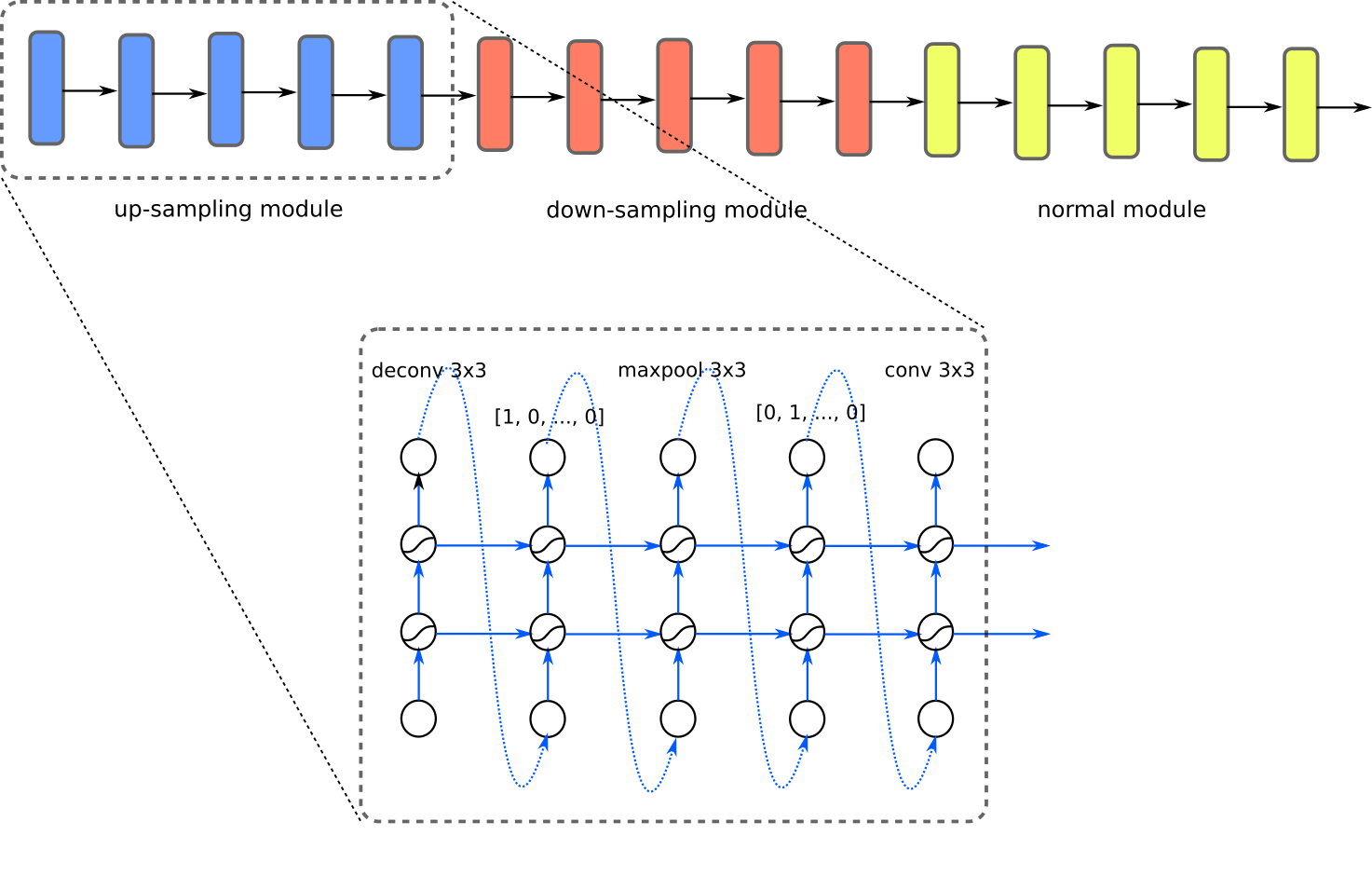}
    \caption{%
    Controller RNN architecture. Above: The controller consists of three segments, programming the up-sampling module in the generator,
    the down-sampling and normal modules in the discriminator, respectively; Below: In each segment, the controller samples an operation and an adjacency vector in turn in an autoregressive manner.
    }
    \label{fig:controller}
\end{figure*}

The controller is a two-layer LSTM consisting of three segments (Figure~\ref{fig:controller}), programming the up-sampling module in the generator, the down-sampling and normal modules in the discriminator, respectively. In each segment, the controller iteratively outputs a candidate operation in the module and an adjacency vector indicating tensors that will be fed into the incoming operation; the output, either an operation or an adjacency vector, is then fed into next step as input.

All operations are sampled through a softmax classifier with sample temperature $T$ and logit clipping constant $C$ \cite{NCO}
\[
    \Pr{y = y_i} \propto \exp{\left( C\tanh{(h_i/T)} \right)},
\]
where $y$ is the output operation and $\Vec{h}=(h_i)$ is the last hidden layer at current time step. $y$ is fed into the controller RNN through an embedding layer; the embedding parameters are only shared within the same segment.

The adjacency vector is sampled from element-wise independent multivariate Bernoulli distribution
\[
    \Pr{\Vec{y} = (y_i)} = \prod_i {p_i}^{y_i} (1 - p_i)^{1 - y_i},
\]
\[
    p_i = \sigma \left( C\tanh{(h_i/T)} \right),
\]
where $\Vec{y}$ is the adjacency vector and $\sigma$ is the sigmoid activation. $\Vec{y}$ is fed into the controller RNN through a linear projection layer whose parameters are similarly shared within the same segment.

\subsection{The Search Space}

The outputs of each segment in the controller RNN will be used to program a module in the child model. At each time step, we select tensors according to the sampled adjacency vector, and feed their sum into next operation.

More precisely, each module takes the output of last two modules, $h_{i-1}$ and $h_i$, as inputs. The output sequence of controller RNN segment $o_0, \Vec{a_0}, o_1, \Vec{a_1}, ..., \Vec{a_{k-1}}, o_k$ always starts with and ends with an operation. The module is constructed as follows:
\begin{step}
    \item Apply the first operation $o_0$ to $h_{i-1}$ to form the skip connection. Let $\mathcal{N} = \{x_0, x_1\}$, where $x_0 = h_i$ and $x_1 = o_0(h_{i-1})$.
    \item For each $i$, select tensors from $\mathcal{N}$ according to $\Vec{a_i}$. If $\Vec{a_i} = \Vec{0}$, input of the module $x_0$ will be selected.
    \item Apply $o_i$ to the sum of selected tensors and add resulting tensor to $\mathcal{N}$.
    \item Repeat Step 2 and Step 3.
    \item Concatenate tensors in $\mathcal{N}$ who have never served as an input to form the final output.
\end{step}

\begin{figure}[h!]
    \centering
    \includegraphics[width=.6\linewidth]{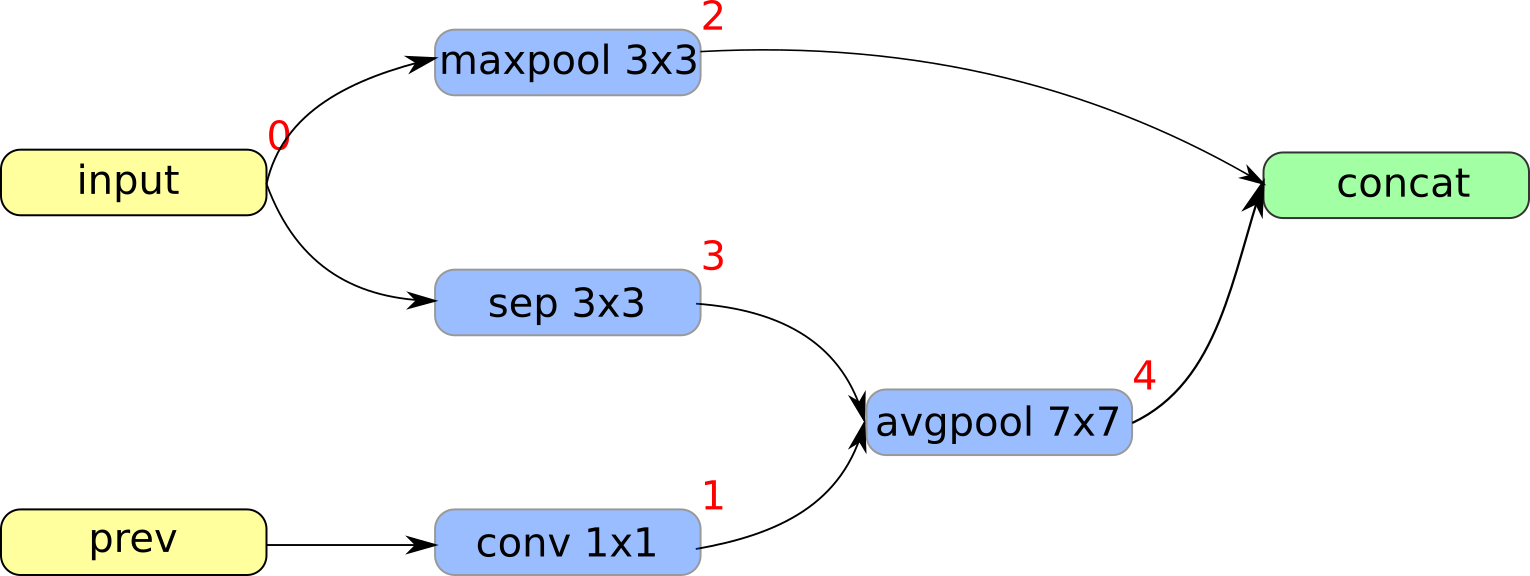}
    \caption{An normal module defined by controller sequence \texttt{conv 1x1}, $(1, 0, 0, 0)$,  \texttt{maxpool 3x3}, $(0, 0, 0, 0)$, \texttt{sep 3x3}, $(0, 1, 0, 1)$, \texttt{avgpool 7x7}. 
    1) Apply \texttt{conv 1x1} to \texttt{prev}. 
    2) Select \texttt{input} according to vector $(1, 0, 0, 0)$ and apply \texttt{maxpool 3x3}. 
    3) No tensors selected. Apply \texttt{sep 3x3} to \texttt{input}. 
    4) Sum over tensor $1$ and $3$ as instructed by $(0, 1, 0, 1)$. Apply \texttt{avgpool 7x7}. 
    5) Concatenate tensor $2$ and $4$ to form the final output.
    }
    \label{fig:topology}
\end{figure}

In addition, we adopt the following heuristics to ensure the computation graph is well-defined:
\begin{itemize}
    \item The first operation is interpreted as an up-sampling (down-sampling) operation if the previous module is an up-sampling (down-sampling) module.
    \item For up-sampling modules, the operations applied to $x_0$, $x_1$ will be interpreted as up-sampling operations.
    \item For down-sampling modules, the operations applied right before concatenation will be interpreted as down-sampling operations.
    \item $1\times1$ convolutions are applied to the final output to keep the number of channels constant.
\end{itemize}

The meta-architectures of the generator and the discriminator are manually determined as follows. Starting with a linear layer, the generator consists of $3$ up-sampling modules, followed by a $3\times3$ convolution and a $\tanh$ activation. The discriminator starts with a $3\times3$ convolution, followed by $2$ down-sampling modules, $2$ normal modules, a global sum pooling layer and a linear layer. For conditional version of the model, the discriminator logits is augmented with a projection layer as in \cite{proj-dis}.

\begin{figure}[h!]
    \centering
    \includegraphics[width=.5\linewidth]{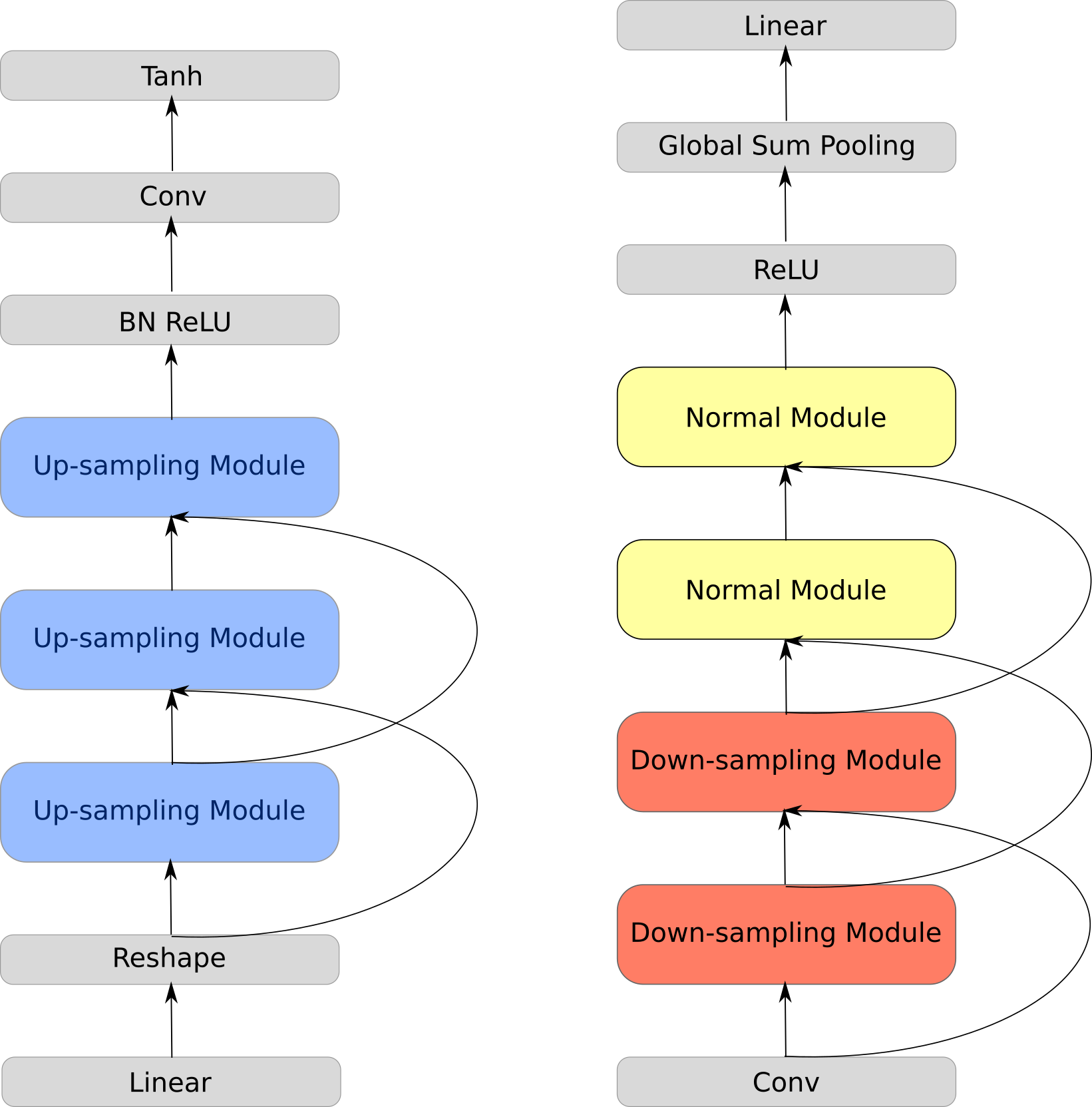}
    \caption{Meta-architecture of the generator and discriminator}
    \label{fig:meta-arch}
\end{figure}

We use hinge loss \cite{hinge-loss} as the objective function, where
\begin{align*}
L_D = & -\Epp{(x, y)\sim p_{\textrm{data}}}{\min{0, -1 + D(x, y)}} \\
      & -\Epp{z\sim p_z, y\sim p_\textrm{data}}{\min{0, -1 - D(G(z), y)}}, \\
L_G = & -\Epp{z\sim p_z, y\sim p_\textrm{data}} D(G(z), y).
\end{align*}

To cover a large variety of candidate architectures, we collect the following set of operations as our normal operations:
\begin{multicols}{2}
\begin{itemize}[topsep=0pt,itemsep=-1ex,partopsep=1ex,parsep=1ex]
    \item identity
    \item $1\times1$ convolution
    \item $3\times3$ convolution
    \item $3\times3$ dilated convolution
    \item $3\times3$ depthwise-separable convolution
    \item $5\times5$ depthwise-separable convolution
    \item $7\times7$ depthwise-separable convolution
    \item $1\times3$ then $3\times1$ convolution
    \item $1\times5$ then $5\times1$ convolution
    \item $1\times7$ then $7\times1$ convolution
    \item $3\times3$ max pooling
    \item $5\times5$ max pooling
    \item $7\times7$ max pooling
    \item $3\times3$ average pooling
    \item $5\times5$ average pooling
    \item $7\times7$ average pooling
\end{itemize}
\end{multicols}

For up-sampling modules, based on state-of-the-art GAN architectures we consider two different types of up-sampling operations: 1) $3\times3$, $5\times5$ or $7\times7$ transposed convolution 2) nearest-neighbor interpolation followed by any convolution in the list above. For down-sampling modules, motivated by optimized residual blocks in \cite{WGAN-GP}, we include 1) convolution followed by stride $2$ average pooling 2) stride $2$ average pooling followed by convolution as two types of  atomic operations.

We use BN - ReLU - Conv for all convolutional operations in G, and ReLU - Conv for all convolutional operations in D. There is no Batch Normalization nor ReLU in between $1\times n$ then $n\times 1$ convolutions. 

\subsection{Training with Reinforcement Learning}

We use REINFORCE rule \cite{reinforce} to udpate controller RNN parameters $\theta$. Let $a_0, a_1, \ldots, a_T$ be the output sequence of controller, including both operational and connectivity choices. We have the following update rule for $\theta$:
\[
    \nabla_\theta J(\theta) = 
    \sum_t \Ep{(R - b) \nabla_\theta \log \Pr{a_t | a_{t-1}}},
\]
where $R$ is the reward for taking actions $a_0, a_1, \ldots, a_T$ and $b$ is the baseline for variance reduction. In particular, when $a_t$ is an operation or adjacency vector, $-\log \Pr{a_t | a_{t-1}}$ can be computed through softmax or sigmoid cross-entropy.

We measure the performance of GAN using Inception Score. More precisely, we propose the following reward shaping
\[
    R = \frac{\IS - \IS_\textrm{min}}{\IS_\textrm{max} - \IS},
\]
where $\IS_\textrm{min}$ and $\IS_\textrm{min}$ are constants, making the rewards more sensitive when IS approaching optimal value. Due to the instability of GAN training, the Inception Score needs to be averaged over multiple run of GAN to ensure reliable measurement. In practice, however, we found that the proposed NAS algorithm works with a single run of training per sampled architecture.
\section{Experiments}{\label{sec:experimentalStudy}}

\subsection{Data Sets}
We used two data sets in our experimental study. The CIFAR-10 data set consists of $60,000$ $32\times32$ color images in $10$ different classes. The data set is divided into a training data set of $50,000$ images and a testing data set with the rest $10,000$ images. Only training set is used for our experiment.  The STL-10 data set is an image data set of $96\times96$ color images. It is composed of images in $10$ different classes with $500$ training images and $800$ testing images per class, and an additional $100,000$ unlabeled images for unsupervised learning. 

For data preprocessing, we follow the setup in \cite{SNGAN} by scaling the images to $[-1, \frac{127}{128}]$ then adding random noise $\epsilon\sim\mathcal{U}(0, \frac{1}{128})$ for both data sets. 

\subsection{Experimental Procedure}

The controller used in our experimental study is a two-layer LSTM with $100$ units, consisting of three segments. Each segment outputs a sequence of $11$ actions ($6$ operations and $5$ adjacency vectors), encoding a DAG of $6$ nodes. We use sample temperature $T = 5.0$ and logit clipping $C = 2.5$ when sampling operations, and $T = C = 1.0$ when sampling adjacency vectors. The controller is trained using policy gradient with learning rate $0.0006$. We compensate the loss with an entropy temperature $0.0001$ to ensure better exploration. The controller is updated whenever $10$ rewards are collected from child models. We use $200$ Titan X GPUs training for $6$ days, with an overall sample complexity of $20,000$.

When constructing the GAN model, we fix the number of channels in both the generator and discriminator to be $128$. We find that using global sum pooling instead of global average pooling in the penultimate layer of the discriminator stabilizes the training. We use Adam optimizer \cite{Adam} for optimization with $\eta = 0.0002$, $\beta_1 = 0$ and $\beta_2 = 0.9$. The discriminator is updated $5$ steps per one generator update step. The number of samples generated is $128$ per G update and $64$ per D update. We use batch size $64$ for real samples. To evaluate the architecture, the model is  trained for $5,000$ steps. Inception Score is then calculated based on $50,000$ generated samples divided into $10$ groups. For reward shaping, we choose $\IS_\textrm{max} = 11.24$ (Inception Score of the real data) and $\IS_\textrm{min} = 1$.

\begin{figure}[H]
    \centering
    \begin{subfigure}[h!]{0.5\linewidth}
        \centering
        \includegraphics[width=\linewidth]{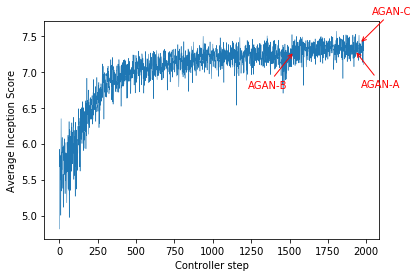}
        \caption{Averge Inception Score}
    \end{subfigure}%
    \begin{subfigure}[h!]{0.5\linewidth}
        \centering
        \includegraphics[width=\linewidth]{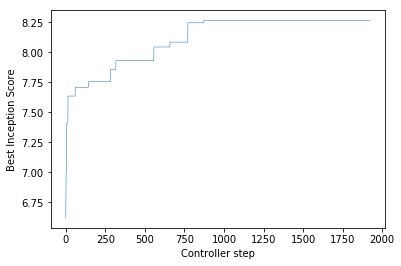}
        \caption{Best Inception Score}
    \end{subfigure}
    \caption{Progression of Inception Score on CIFAR-10 \\
    Model with the highest Inception Score (when trained for $5,000$ steps) is generated as early as step $=913$. The controller, however, continues to learn the distribution that samples better performing models on average. In fact, the best models when trained to full size are generated at the later stage of the architecture search.
    }
    \label{fig:converge}
\end{figure}

\subsection{Learning GAN architecture on CIFAR-10}

For the task of supervised image generation on CIFAR-10, we take top candidate models discovered in the architecture search and train for $50,000$ steps. We scale up the models by doubling the number of channels in both the generator and the discriminator. The label information is fed into G via Conditional Batch Normalization (CBN) \cite{cond-batch-norm} and into D via projection \cite{proj-dis}. We use Spectral Normalization \cite{SNGAN} for the discriminator but not the generator. The best architectures are reported in Table~\ref{tab:supervised}.

\begin{table}[h!]
\centering
\begin{tabular}{p{1.5in}ll}
\hline
Method & \multicolumn{1}{c}{Inception Score} & \multicolumn{1}{c}{FID} \\
\hline
Real data & $11.24\pm.12$ & $7.8$ \\
\hline
\textbf{DCGAN style} \\
SteinGAN & $6.35$ \\
DCGAN & $6.58$ \cite{learn2draw} \\
Salimans et al. \cite{Inception-score} & $8.09\pm.07$ \\
AC-GAN & $8.25\pm.07$ \\
SGAN & $8.59\pm.12$ \\
\hline
\textbf{ResNet} \\
WGAN-GP & $8.42\pm.12$ & \\
SN-GAN & $8.62$ \cite{proj-dis} & $17.5$ \cite{proj-dis} \\
BigGAN & $\bm{9.22}$  & $\bm{14.73}$ \\
\hline
\textbf{Ours} \\
\name-A & $8.65\pm.12$ & $23.0$ \\
\name-B & $8.82\pm.09$ & $23.8$ \\
\name-C & $8.55\pm.11$ & $28.0$ \\
\end{tabular}
\caption{Supervised image generation on CIFAR-10}
\label{tab:supervised}
\end{table}

\begin{figure}[h!]
    \centering
    \begin{subfigure}[h!]{.3\linewidth}
    \includegraphics[width=\linewidth]{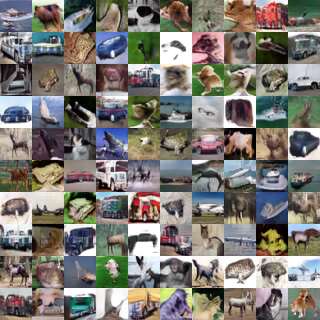}
    \caption{\name-A}
    \end{subfigure}%
    ~
    \begin{subfigure}[h!]{.3\linewidth}
    \includegraphics[width=\linewidth]{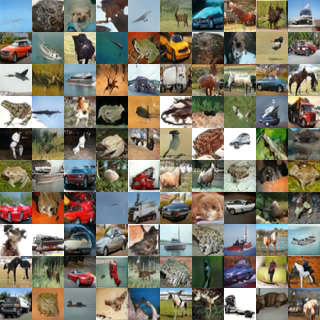}
    \caption{\name-B}
    \end{subfigure}%
    ~
    \begin{subfigure}[h!]{.3\linewidth}
    \includegraphics[width=\linewidth]{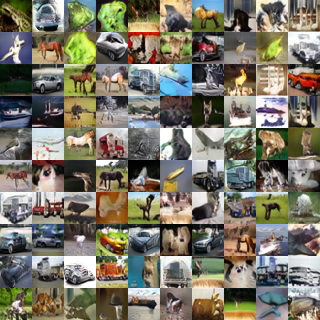}
    \caption{\name-C}
    \end{subfigure}
    \caption{Images generated by \name s in supervised image generations tasks}
    \label{fig:supervised-image}
\end{figure}
\name-A and \name-B outperform all DCGAN-style architectures. The best architecture we found, AGAN-B, also outperforms \emph{all} ResNet architectures with $32\times32$ input resolution or less than $100$M parameters. In particular, BigGAN \cite{BigGAN} architecture resides on $128\times128$ input images and has $158.3$ parameters. The architectures we proposed have much less parameters ($20.1$M, $18.8$M and $20.9$M, respectively) in comparison.

We also train models with the same topology for unsupervised image generation tasks. We drop the projection layer in D and use Batch Normalization in place of CBN in G. We find that scaling up does not guarantee performance gain  in this setting. All of the architectures proposed outperform DCGAN-style architectures and AGAN-C outperforms all ResNet architectures in terms of Inception Score.

\begin{table}[h!]
\centering
\begin{tabular}{p{1.5in}ll}
\hline
Method & \multicolumn{1}{c}{Inception Score} & \multicolumn{1}{c}{FID} \\
\hline
Real data & $11.24\pm.12$ & $7.8$ \\
\hline
\textbf{DCGAN style} \\
BEGAN & $5.62$ \\
DCGAN & $6.16\pm.07$ \cite{SGAN} & $36.9$ \cite{FID}\\
MMD GAN & $6.17\pm.07$ & $38.2$ \cite{Coulomb-GANs} \\
WGAN-GP & $6.68\pm.06$ \cite{SNGAN} & 24.8 \cite{FID} \\
Salimans et al. & $6.86\pm.06$ \\
LR-GAN & $7.17\pm.07$ \\
SN-GAN & $7.42\pm.08$ & $29.3$ \\
DFM & $7.72\pm.13$ \\
Coulomb GANs & & 27.3 \\
\hline
\textbf{ResNet} \\
WGAN-GP & $7.86\pm.07$ &\\
SN-GAN & $8.22\pm.05$ & $\bm{21.7\pm.21}$\\
\hline
\textbf{Ours} \\
\name-A & $8.15\pm.09$ & $31.6$ \\
\name-B & $7.77\pm.10$ & $33.2$ \\
\name-C & $\bm{8.29\pm.09}$ & $30.5$ \\
\end{tabular}
\caption{Unsupervised image generation on CIFAR-10}
\label{tab:unsupervised}
\end{table}

\begin{figure}[h!]
    \centering
    \begin{subfigure}[h!]{.5\linewidth}
        \includegraphics[width=\linewidth]{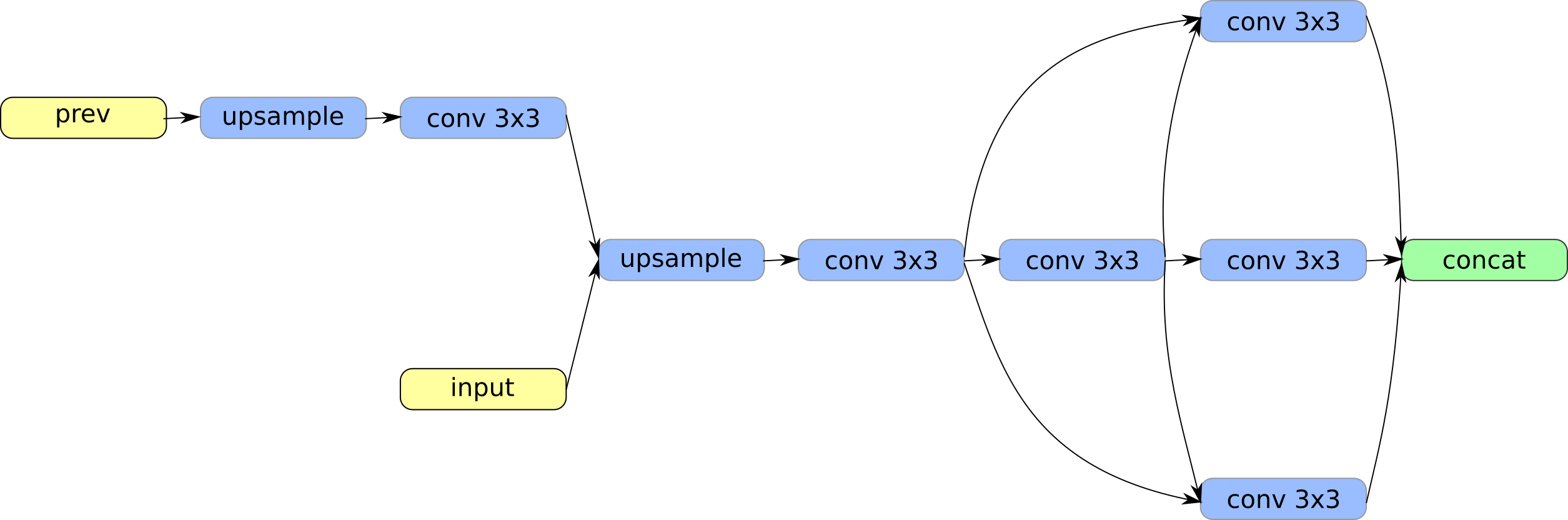}
        \caption{Up-sampling module}
    \end{subfigure}
    \begin{subfigure}[h!]{0.45\linewidth}
        \includegraphics[width=\linewidth]{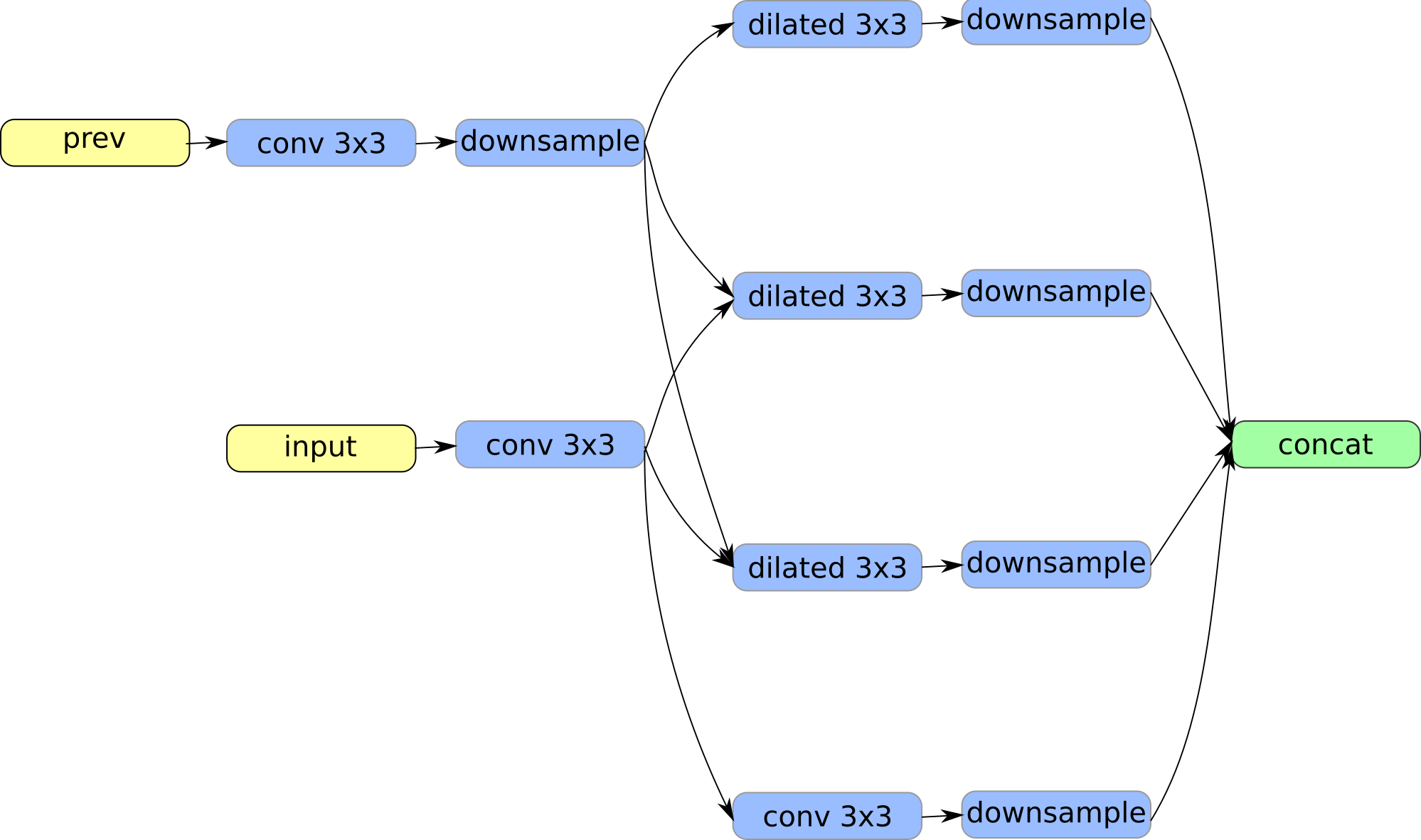}
        \caption{Down-sampling module}
    \end{subfigure}
    \begin{subfigure}[h!]{0.475\linewidth}
        \includegraphics[width=\linewidth]{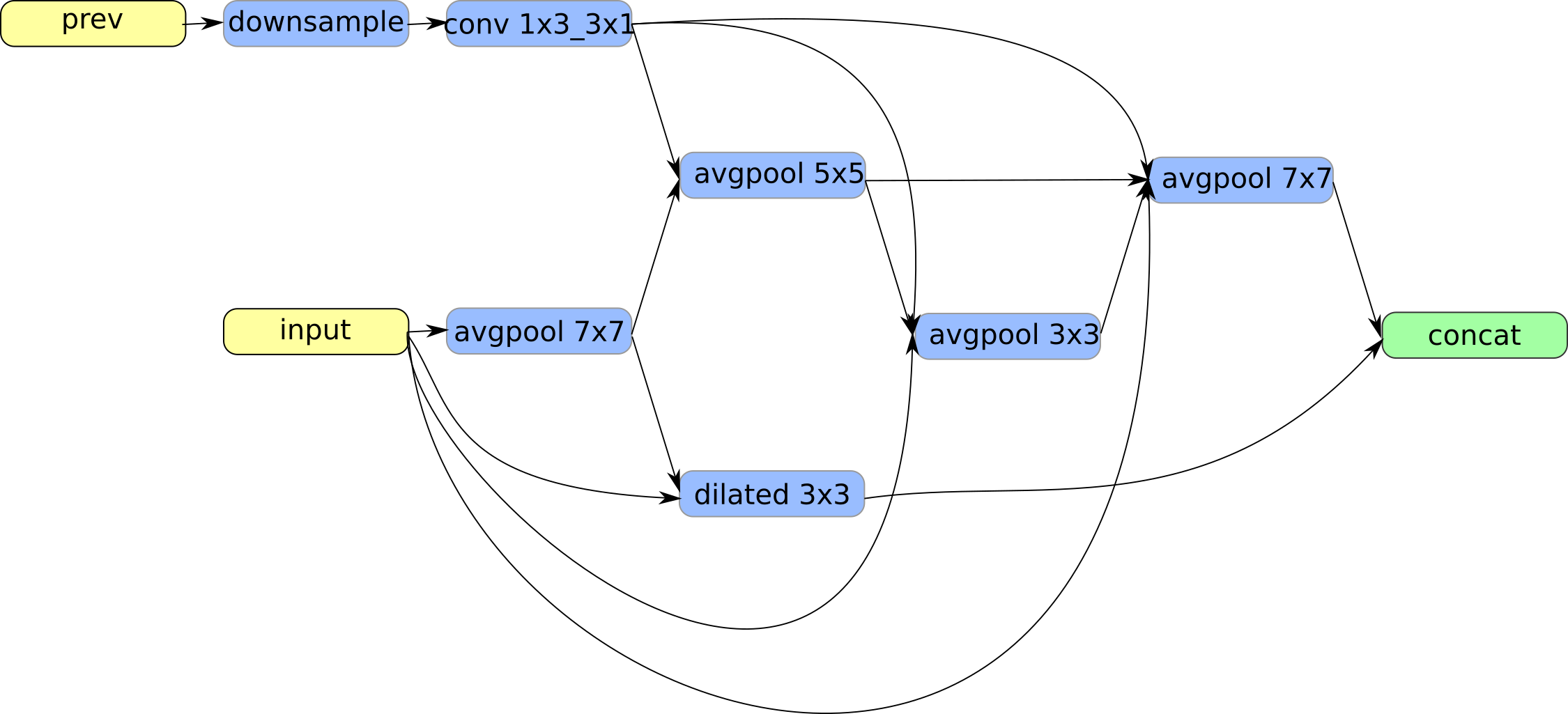}
        \caption{Normal module}
    \end{subfigure}
    \caption{%
    Topology of modules in AGAN-A \\
    Note that the up-sample (down-sample) operations following \texttt{prev} will only be applied when the module is preceded by an up-sampling (down-sampling) module.
    }
    \label{fig:agan-a}
\end{figure}

In Figure \ref{fig:agan-a} we decipher the architecture of the learned model \name-A. Note that topology of all three modules: up-sampling, down-sampling, and normal ones, are quite different from modules used in existing models. Such architecture is a hybrid between Inception and Resnet in that each cell, as deciphered here, contains multiple branches. Cells are stacked together in a way resembling Resnet as shown previously in Figure \ref{fig:meta-arch}. We believe this is the first time that we see inception-resnet hybrid architectures that are used for GAN. Also the cells that we see here are quite different from inception cells that we typically use in discriminative models, which provide evidence supporting our original idea that optimal GAN architecture could be quite different from those in discriminative models. Architectures of \name-B and \name-C bear some resemblance to \name-A and we omit their diagrams for brevity. 

\subsection{Transferability of \name}
One potential advantage of modularzied search space is that it enables the transferability of the learned architecture: modules generated on smaller data sets could be used as building blocks to construct networks on larger data sets, where direct neural architecture search may be infeasible or unfavorable. In this experiment, we empirically evaluated the transferability of some of our learned modules, namely \name-A and \name-C. 

Our STL-10 network has the same meta-architecture as the one for CIFAR-10, with the distinction that the first up-sampling module in G takes input size of $6\times6\times n$ (instead of $4\times4\times n$). We resize the STL-10 data set to $48\times48$ images. As in Table~\ref{tab:stl10}, despite that their topology are not optimized for STL-10, \name-A and \name-C achieve highly competitive performances, outperforming all DCGAN-style architectures. The experiment provide evidences suggesting that the architectures that we identified might be applicable to a wide range of data sets. 

\begin{table}[H]
\centering
\begin{tabular}{p{1.5in}ll}
\hline
Method & \multicolumn{1}{c}{Inception Score} & \multicolumn{1}{c}{FID} \\
\hline
Real data & $11.24\pm.12$ & $7.8$ \\
\hline
\textbf{DCGAN style} \\
DCGAN & $7.84\pm0.07$ \cite{DFM} & \\
DFM & $8.51\pm0.13$\\
SN-GAN & $8.69\pm.09$ & $47.5$ \\
\hline
\textbf{ResNet} \\
WGAN-GP & $9.05\pm.13$ \cite{Splitting-GAN} & $55.1$ \cite{SNGAN} \\
SN-GAN & $9.10\pm.04$ & $\bm{40.1}$ \\
Splitting GAN & $9.50\pm.13$ \\
CAGAN & $\bm{9.51\pm.14}$ \\
\hline
\textbf{Ours} \\
\name-A & $9.23\pm.08$ & $52.7$ \\
\name-B & $7.84\pm.12$ & $71.8$ \\
\name-C & $8.97\pm.10$ & $57.4$ \\
\end{tabular}
\caption{Unsupervised image generation on STL-10}
\label{tab:stl10}
\end{table}

\begin{figure*}[h!]
    \centering
    \begin{subfigure}[h!]{.3\linewidth}
    \includegraphics[width=\linewidth]{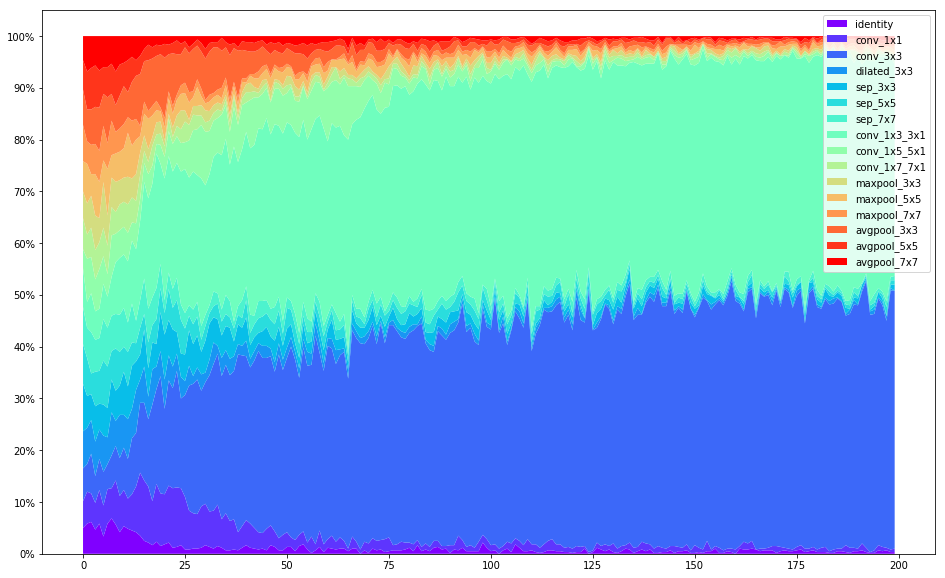}
    \caption{Up-sampling}
    \end{subfigure}%
    ~
    \begin{subfigure}[h!]{.3\linewidth}
    \includegraphics[width=\linewidth]{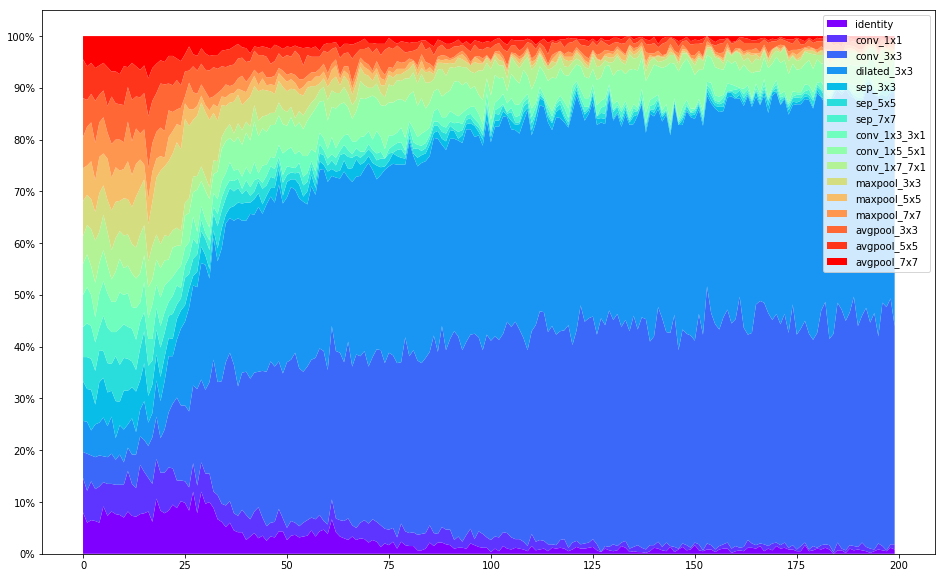}
    \caption{Down-sampling}
    \end{subfigure}%
    ~
    \begin{subfigure}[h!]{.3\linewidth}
    \includegraphics[width=\linewidth]{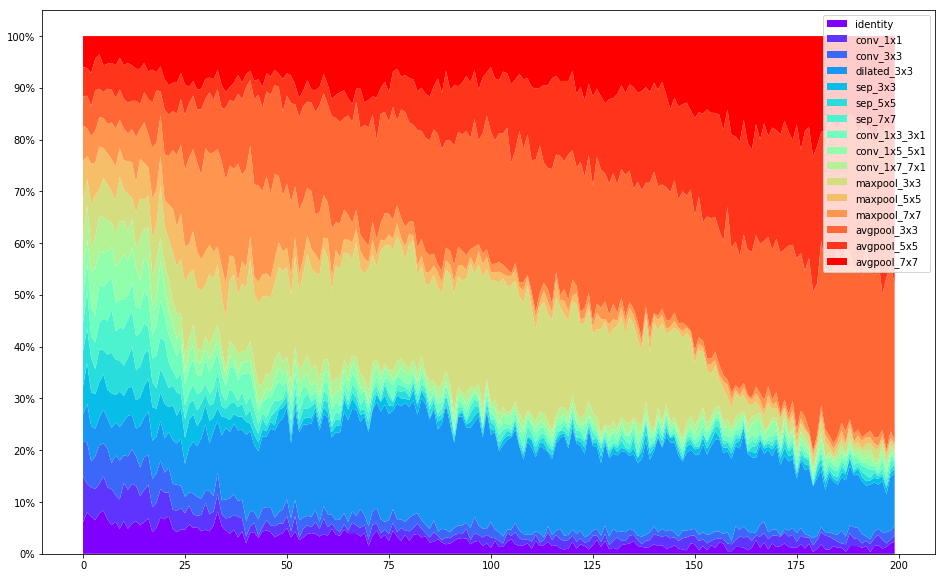}
    \caption{Normal}
    \end{subfigure}
    \caption{%
    Empirical distribution of sampled operations by module over time \\
    Learned operations: (a) up-sampling module: conv $3\times3$, conv $1\times3$ then $3\times1$ (b) down-sampling module: conv $3\times3$, dilated conv $3\times3$ (c) normal module: average pooling $3\times3$, $5\times5$, $7\times7$
    }
    \label{fig:learn-ops}
\end{figure*}
\section{Discussion \& Conclusion}{\label{sec:conclusion}}

As illustrated in Figure~\ref{fig:learn-ops}, in our search of GAN architectures, the controller RNN learns drastically different distributions over operations for three module types. The up-sampling modules predominately favor conv $3\times3$ and conv $1\times3$ then $3\times1$; the down-sampling modules favor conv $3\times3$ and dilated conv $3\times3$; the normal modules favor average pooling $3\times3$, $5\times5$ and $7\times7$. This justifies our choice of segmentation of controller RNN. We point out that it is at least in contrast to RL-based NAS algorithms over image classifiers \cite{ENAS, BlockQNN,NAS2}, where both the normal cell and reduction cell choose among depthwise-separable convolutions, max and average poolings.

In addition, we observe that
\begin{itemize}
    \item the up-sampling modules prefer upsample-then-convolution operations, over transposed convolutions;
    \item the down-sampling modules prefer convolution-then-downsample operations over downsample-then-convolutions;
    \item the normal modules mostly consist of average poolings, and hence have very few parameters;
    \item depthwise-separable convolutions are not present at most networks in later stage.
\end{itemize}

In our experiments we observe that the order of operations in the same module matters much. For example, in down-sampling modules, whether we perform down-sampling at the beginning of the module or at the end of the model may have significant impact on the overall performance though the exact mechanism of the effect is not clear. 

In conclusion, we present \name, the first neural architecture search algorithm on deep generative models. We demonstrate that, by careful design of controller architecture and search space, RL-based NAS algorithm can discover highly competitive architectures that rival the best human-invented GAN architecture. Further reducing model size and enabling fast inference are on our future research agenda.

\bibliographystyle{unsrt}  
\bibliography{egbib}  

\begin{thebibliography}{10}

\bibitem{GAN}
Ian Goodfellow, Jean Pouget-Abadie, Mehdi Mirza, Bing Xu, David Warde-Farley,
  Sherjil Ozair, Aaron Courville, and Yoshua Bengio.
\newblock Generative adversarial nets.
\newblock In Z.~Ghahramani, M.~Welling, C.~Cortes, N.~D. Lawrence, and K.~Q.
  Weinberger, editors, {\em Advances in Neural Information Processing Systems
  27}, pages 2672--2680. Curran Associates, Inc., 2014.

\bibitem{super-resolution}
Christian Ledig, Lucas Theis, Ferenc Huszar, Jose Caballero, Andrew~P. Aitken,
  Alykhan Tejani, Johannes Totz, Zehan Wang, and Wenzhe Shi.
\newblock Photo-realistic single image super-resolution using a generative
  adversarial network.
\newblock {\em CoRR}, abs/1609.04802, 2016.

\bibitem{text-to-image}
Scott~E. Reed, Zeynep Akata, Xinchen Yan, Lajanugen Logeswaran, Bernt Schiele,
  and Honglak Lee.
\newblock Generative adversarial text to image synthesis.
\newblock {\em CoRR}, abs/1605.05396, 2016.

\bibitem{CycleGAN}
Jun{-}Yan Zhu, Taesung Park, Phillip Isola, and Alexei~A. Efros.
\newblock Unpaired image-to-image translation using cycle-consistent
  adversarial networks.
\newblock {\em CoRR}, abs/1703.10593, 2017.

\bibitem{WGAN}
Martin Arjovsky, Soumith Chintala, and L{\'e}on Bottou.
\newblock {W}asserstein generative adversarial networks.
\newblock In Doina Precup and Yee~Whye Teh, editors, {\em Proceedings of the
  34th International Conference on Machine Learning}, volume~70 of {\em
  Proceedings of Machine Learning Research}, pages 214--223, International
  Convention Centre, Sydney, Australia, 06--11 Aug 2017. PMLR.

\bibitem{LS-GAN}
Xudong Mao, Qing Li, Haoran Xie, Raymond Y.~K. Lau, and Zhen Wang.
\newblock Multi-class generative adversarial networks with the {L2} loss
  function.
\newblock {\em CoRR}, abs/1611.04076, 2016.

\bibitem{f-GAN}
Sebastian Nowozin, Botond Cseke, and Ryota Tomioka.
\newblock f-gan: Training generative neural samplers using variational
  divergence minimization.
\newblock In D.~D. Lee, M.~Sugiyama, U.~V. Luxburg, I.~Guyon, and R.~Garnett,
  editors, {\em Advances in Neural Information Processing Systems 29}, pages
  271--279. Curran Associates, Inc., 2016.

\bibitem{generator-conditioning}
Augustus Odena, Jacob Buckman, Catherine Olsson, Tom Brown, Christopher Olah,
  Colin Raffel, and Ian Goodfellow.
\newblock Is generator conditioning causally related to {GAN} performance?
\newblock In Jennifer Dy and Andreas Krause, editors, {\em Proceedings of the
  35th International Conference on Machine Learning}, volume~80 of {\em
  Proceedings of Machine Learning Research}, pages 3849--3858,
  Stockholmsmässan, Stockholm Sweden, 10--15 Jul 2018. PMLR.

\bibitem{WGAN-GP}
Ishaan Gulrajani, Faruk Ahmed, Mart{\'{\i}}n Arjovsky, Vincent Dumoulin, and
  Aaron~C. Courville.
\newblock Improved training of wasserstein gans.
\newblock {\em CoRR}, abs/1704.00028, 2017.

\bibitem{SNGAN}
Takeru Miyato, Toshiki Kataoka, Masanori Koyama, and Yuichi Yoshida.
\newblock Spectral normalization for generative adversarial networks.
\newblock {\em CoRR}, abs/1802.05957, 2018.

\bibitem{DCGAN}
Alec Radford, Luke Metz, and Soumith Chintala.
\newblock Unsupervised representation learning with deep convolutional
  generative adversarial networks.
\newblock {\em CoRR}, abs/1511.06434, 2015.

\bibitem{NAS1}
Barret Zoph and Quoc~V. Le.
\newblock Neural architecture search with reinforcement learning.
\newblock {\em CoRR}, abs/1611.01578, 2016.

\bibitem{NAS2}
Barret Zoph, Vijay Vasudevan, Jonathon Shlens, and Quoc~V. Le.
\newblock Learning transferable architectures for scalable image recognition.
\newblock {\em CoRR}, abs/1707.07012, 2017.

\bibitem{regularized-evolution}
Esteban Real, Alok Aggarwal, Yanping Huang, and Quoc~V. Le.
\newblock Regularized evolution for image classifier architecture search.
\newblock {\em CoRR}, abs/1802.01548, 2018.

\bibitem{SAGAN}
Han Zhang, Ian~J. Goodfellow, Dimitris~N. Metaxas, and Augustus Odena.
\newblock Self-attention generative adversarial networks.
\newblock {\em arXiv:1805.08318}, 2018.

\bibitem{BigGAN}
Andrew Brock, Jeff Donahue, and Karen Simonyan.
\newblock Large scale {GAN} training for high fidelity natural image synthesis.
\newblock {\em CoRR}, abs/1809.11096, 2018.

\bibitem{cGAN}
Mehdi Mirza and Simon Osindero.
\newblock Conditional generative adversarial nets.
\newblock {\em CoRR}, abs/1411.1784, 2014.

\bibitem{proj-dis}
Takeru Miyato and Masanori Koyama.
\newblock cgans with projection discriminator.
\newblock {\em CoRR}, abs/1802.05637, 2018.

\bibitem{cond-batch-norm}
Harm de~Vries, Florian Strub, J{\'{e}}r{\'{e}}mie Mary, Hugo Larochelle,
  Olivier Pietquin, and Aaron~C. Courville.
\newblock Modulating early visual processing by language.
\newblock {\em CoRR}, abs/1707.00683, 2017.

\bibitem{Theis}
Lucas Theis, A{\"a}ron van~den Oord, and Matthias Bethge.
\newblock A note on the evaluation of generative models.
\newblock {\em CoRR}, abs/1511.01844, 2016.

\bibitem{Inception-score}
Tim Salimans, Ian~J. Goodfellow, Wojciech Zaremba, Vicki Cheung, Alec Radford,
  and Xi~Chen.
\newblock Improved techniques for training gans.
\newblock {\em CoRR}, abs/1606.03498, 2016.

\bibitem{Google-Inception}
Christian Szegedy, Wei Liu, Yangqing Jia, Pierre Sermanet, Scott~E. Reed,
  Dragomir Anguelov, Dumitru Erhan, Vincent Vanhoucke, and Andrew Rabinovich.
\newblock Going deeper with convolutions.
\newblock {\em CoRR}, abs/1409.4842, 2014.

\bibitem{EAS}
Han Cai, Tianyao Chen, Weinan Zhang, Yong Yu, and Jun Wang.
\newblock Efficient architecture search by network transformation.
\newblock In {\em AAAI}, 2018.

\bibitem{path-level-EAS}
Han Cai, Jiacheng Yang, Weinan Zhang, Song Han, and Yong Yu.
\newblock Path-level network transformation for efficient architecture search.
\newblock {\em arXiv preprint arXiv:1806.02639}, 2018.

\bibitem{ENAS}
Hieu Pham, Melody~Y. Guan, Barret Zoph, Quoc~V. Le, and Jeff Dean.
\newblock Efficient neural architecture search via parameter sharing.
\newblock {\em CoRR}, abs/1802.03268, 2018.

\bibitem{MetaQNN}
Bowen Baker, Otkrist Gupta, Nikhil Naik, and Ramesh Raskar.
\newblock Designing neural network architectures using reinforcement learning.
\newblock {\em CoRR}, abs/1611.02167, 2016.

\bibitem{BlockQNN}
Zhao Zhong, Zichen Yang, Boyang Deng, Junjie Yan, Wei Wu, Jing Shao, and
  Cheng{-}Lin Liu.
\newblock Blockqnn: Efficient block-wise neural network architecture
  generation.
\newblock {\em CoRR}, abs/1808.05584, 2018.

\bibitem{evolution}
Esteban Real, Sherry Moore, Andrew Selle, Saurabh Saxena, Yutaka~Leon Suematsu,
  Quoc~V. Le, and Alex Kurakin.
\newblock Large-scale evolution of image classifiers.
\newblock {\em CoRR}, abs/1703.01041, 2017.

\bibitem{hierarchical-CNN}
Hanxiao Liu, Karen Simonyan, Oriol Vinyals, Chrisantha Fernando, and Koray
  Kavukcuoglu.
\newblock Hierarchical representations for efficient architecture search.
\newblock {\em CoRR}, abs/1711.00436, 2017.

\bibitem{DARTS}
Hanxiao Liu, Karen Simonyan, and Yiming Yang.
\newblock {DARTS:} differentiable architecture search.
\newblock {\em CoRR}, abs/1806.09055, 2018.

\bibitem{NAO}
Renqian Luo, Fei Tian, Tao Qin, and Tie{-}Yan Liu.
\newblock Neural architecture optimization.
\newblock {\em CoRR}, abs/1808.07233, 2018.

\bibitem{proxyless-NAS}
Han Cai, Ligeng Zhu, and Song Han.
\newblock Proxylessnas: Direct neural architecture search on target task and
  hardware.
\newblock {\em CoRR}, abs/1812.00332, 2018.

\bibitem{PNAS}
Chenxi Liu, Barret Zoph, Jonathon Shlens, Wei Hua, Li{-}Jia Li, Li~Fei{-}Fei,
  Alan~L. Yuille, Jonathan Huang, and Kevin Murphy.
\newblock Progressive neural architecture search.
\newblock {\em CoRR}, abs/1712.00559, 2017.

\bibitem{reinforce}
Ronald~J. Williams.
\newblock Simple statistical gradient-following algorithms for connectionist
  reinforcement learning.
\newblock In {\em Machine Learning}, pages 229--256, 1992.

\bibitem{NCO}
Irwan Bello, Hieu Pham, Quoc~V. Le, Mohammad Norouzi, and Samy Bengio.
\newblock Neural combinatorial optimization with reinforcement learning.
\newblock {\em CoRR}, abs/1611.09940, 2016.

\bibitem{hinge-loss}
Jae~Hyun Lim and Jong~Chul Ye.
\newblock Geomtric gan.
\newblock {\em arXiv preprint arXiv:1705.02894}, 2017.

\bibitem{Adam}
Diederik~P. Kingma and Jimmy Ba.
\newblock Adam: {A} method for stochastic optimization.
\newblock {\em CoRR}, abs/1412.6980, 2014.

\bibitem{learn2draw}
Yihao Feng, Dilin Wang, and Qiang Liu.
\newblock Learning to draw samples with amortized stein variational gradient
  descent.
\newblock In {\em Proceedings of the Thirty-Third Conference on Uncertainty in
  Artificial Intelligence, {UAI} 2017, Sydney, Australia, August 11-15, 2017},
  2017.

\bibitem{SGAN}
Xun Huang, Yixuan Li, Omid Poursaeed, John~E. Hopcroft, and Serge~J. Belongie.
\newblock Stacked generative adversarial networks.
\newblock {\em CoRR}, abs/1612.04357, 2016.

\bibitem{FID}
Martin Heusel, Hubert Ramsauer, Thomas Unterthiner, Bernhard Nessler,
  G{\"{u}}nter Klambauer, and Sepp Hochreiter.
\newblock Gans trained by a two time-scale update rule converge to a nash
  equilibrium.
\newblock {\em CoRR}, abs/1706.08500, 2017.

\bibitem{Coulomb-GANs}
Thomas Unterthiner, Bernhard Nessler, G{\"{u}}nter Klambauer, Martin Heusel,
  Hubert Ramsauer, and Sepp Hochreiter.
\newblock Coulomb gans: Provably optimal nash equilibria via potential fields.
\newblock {\em CoRR}, abs/1708.08819, 2017.

\bibitem{DFM}
David Warde-Farley and Yoshua Bengio.
\newblock Improving generative adversarial networks with denoising feature
  matching.
\newblock In {\em ICLR}, 2017.

\bibitem{Splitting-GAN}
Guillermo~L. Grinblat, Lucas~C. Uzal, and Pablo~M. Granitto.
\newblock Class-splitting generative adversarial networks.
\newblock {\em CoRR}, abs/1709.07359, 2017.

\end{thebibliography}






\end{document}